# Physics Driven Deep Retinex Fusion for Adaptive Infrared and Visible Image Fusion

Yuanjie Gu,[1,2,3] Zhibo Xiao,[1,2,3] Yinghan Guan,[1,2,3,4] Haoran Dai,[2,5] Cheng Liu,[6] Liang Xue,[7,*] Shouyu Wang[1,8,*]

[1]OptiX+ Laboratory, School of Electronics and Information Engineering, Wuxi University, Wuxi, Jiangsu 214105, China

[2]HorizonFlow Laboratory, China

[3]Computational Optics Laboratory, Jiangnan University, Wuxi, Jiangsu 214122, China

[4]School of Electronics and Information Engineering, Three Gorges University, Chongqing, 404020, China

[5]Department of Computer Science, Metropolitan College, Boston University, Boston, Massachusetts, 02215, USA

[6]Shanghai Institute of Optics and Fine Mechanics, Chinese Academy of Sciences, Shanghai 201800, China

[7]College of Electronics and Information Engineering, Shanghai University of Electric Power, Shanghai 200090, China

[8]Single Molecule Nanobiology Laboratory (Sinmolab), Nanjing Agricultural University, Nanjing, Jiangsu, 210095, China

[*]Corresponding authors: shouyu29@cwxu.edu.cn (S.W.) & xueliangokay@gmail.com (L.X.)

**Abstract:** Infrared (IR) imaging can highlight thermal radiation objects even under poor lighting or severe sheltering but suffers from low resolution, contrast, and signal-to-noise ratio. While visible (VIS) light imaging can guarantee abundant texture details of targets, it is invalid in low lighting or sheltering conditions. Therefore, IR and VIS image fusion has more extensive applications, but it is a still challenging work because conventional methods cannot balance dynamic range, edge enhancement, and lightness constancy during fusion. To overcome these drawbacks, we propose a novel self-supervised dataset-free method for adaptive IR and VIS image fusion named Deep Retinex Fusion (DRF). The key idea of DRF is first generating component priors that are disentangled from a physical model using generative networks; then combining these priors, which are captured by networks via adaptive fusion loss functions based on Retinex theory; and finally reconstructing the IR and VIS fusion results. Furthermore, to verify the effectiveness of our reported physics driven DRF, qualitative and quantitative experiments via comparing with other state-of-the-art methods are performed using public datasets and in practical applications. These results prove that DRF can provide distinctions between day and night scenes and preserve abundant texture details and high-contrast IR information. Additionally, DRF can adaptively balance IR and VIS information and has good noise immunity. Therefore, compared to large dataset trained methods, DRF, which works without any dataset, achieves the best fusion performance. The DRF codes are open source and can be found at https://github.com/GuYuanjie/Deep-Retinex-fusion.

# 1. Introduction

IR imaging focuses on wavelengths between 8-14 μm [1] indicating heat radiated from objects, therefore, it can highlight thermal radiation objects even under poor lighting or severe sheltering in many applications, such as in military [2], safety inspection [3], non-destructive testing [4], unmanned driving [5], and so on [6-8]. However, IR imaging suffers from low resolution, contrast, and signal-to-noise ratio. Compared to IR imaging, VIS imaging focusing on wavelengths between 380-760 nm can capture reflected light, guaranteeing abundant texture details of targets. But VIS imaging is invalid in low lighting or sheltering conditions. Considering the advantages of both imaging modes, IR and VIS image fusion has more extensive applications. While it is still a challenging work because conventional fusion methods cannot balance the dynamic range, edge enhancement, and lightness constancy during IR and VIS image fusion. Deep learning can partially solve the above problems to some extent. However, different from other tasks such as super-resolution, deblurring and denoising, it is impossible to obtain a real fused image directly. In other words, the ground truth (label) as the essential element in supervised deep learning cannot be obtained. As a result, many deep image fusion works are self-supervised (unsupervised) and roughly divided into two categories with [9-11] and without [12, 13] a training phase. Among those with a training phase, most of them are CNN-based [9, 11, 14] and GAN-based [10, 15, 16]. For CNN-based methods not having end-to-end frameworks, they use CNNs to extract deep features while often adopt conventional fusion methods, which relying on complicated rules for fusion. For CNN-based methods having end-to-end frameworks, they focus on separate design on loss functions or networks, but inevitably leading to incomprehensive performance. Additionally, most GAN methods force the networks to generate the fused image like both IR and VIS source images, while the fusion ratio between IR and VIS images is not equal or even linear. Besides, for methods without a training phase, they first use pre-trained models to extract multi-level deep features and then reconstruct fused images from extracted features. However, as most pre-trained models are trained for various tasks, they are not suitable for specific IR and VIS image fusion.

To overcome these drawbacks in IR and VIS image fusion, we propose a self-supervised disentangled learning method based on Retinex theory [17] named Deep Retinex Fusion (DRF). According to the Retinex theory, the captured image $I(x,y)$ results from the interaction between the light source $L(x,y)$ and the object $R(x,y)$, expressed as $I(x,y) = R(x,y) \times L(x,y)$. Retinex based on lightness consistency is different from conventional linear and nonlinear methods (like power-law function, gamma function, and histogram equalization), which can only enhance one type of image feature. It can balance dynamic range compression, edge enhancement, and lightness constancy, thus enabling adaptive image enhancement. Inspired by the Retinex theory, physics driven DRF can obtain excellent adaptive IR and VIS image fusion performance without any dataset. DRF captures component priors disentangled by the physical Retinex model combined with our designed dual-path feature switching and skipping network named ZipperNet and adaptive fusion loss functions. In detail, ZipperNet, a generative feature fusion network based on an autoencoder, can fuse deep features with delay feature switching and obtain high generative quality with skipping connections. Meanwhile, the light (radiation) source $L(x,y)$ is generated by our designed single-path network named LightenNet, and the adjusting parameters $\alpha_1$ and $\alpha_2$ are obtained by our designed single-path network named AdjustingNet very similar to LightenNet. Additionally, the adaptive fusion loss functions based on Retinex theory can balance dynamic range compression, edge enhancement, and lightness constancy. Furthermore, compared to other state-of-the-art methods in public datasets and practical applications, DRF achieves the best fusion performance; and more importantly, DRF can adaptively balance IR and VIS information and has good noise immunity.

This paper is organized as follows: Section 2 introduces the physics driven DRF, network architecture, and loss functions. Section 3 compares our proposed DRF with 6 state-of-the-art methods on public datasets, and Section 4 compares them in practical applications. In Section 5, we discussed the DRF effectiveness, and in Section 6, we concluded our study.

## 2. Principle

In this section, we demonstrate the framework, architectures, and adaptive fusion loss functions of the physics driven DRF for IR and VIS image fusion.

*2.1 Framework*

Based on Retinex theory [17], the real world is colorless, and the color perceived is the result of interaction between light and an object, as shown in Fig. 1(A). In other words, the object's color is determined by the object's reflection response to light spectra rather than by the absolute value of the reflected light intensity. The object's lightness is consistent and not affected by the illumination non-uniformity. Different from conventional linear and nonlinear methods used for single-type image feature enhancement, such as addition, gamma and exponential transforms. Retinex satisfying lightness consistency can balance dynamic range compression, edge enhancement, and color constancy, thus supporting the adaptive enhancement of various types of image features. Therefore, Retinex can significantly improve the inconsistency of IR and VIS fusion. In this work, based on the Retinex theory, the recorded image $I(x,y)$ can be disentangled into the reflection image $R(x,y)$ and the lighting image $L(x,y)$, and the fused image can be considered as the reflection image $R(x,y)$. Therefore, we formulate the IR and VIS image fusion as a conditional generative model by ZipperNet, LightingNet, and AdjustingNet. ZipperNet generates the fused image (reflection image $R(x,y)$), and LightingNet generates the transform maps (lighting image $L(x,y)$), as shown in Fig. 1(B). To adjust the overall lightness of the generated fused image, AdjustingNet regresses two parameters $α_1$ and $α_2$. Furthermore, the Retinex loss constructs the relationship among the input images $I(x,y)$, fused image $R(x,y)$, transform maps $L(x,y)$, and adjusting parameters $α_1$ and $α_2$. Significantly, our designed DRF is a non-data-driven method, thus its inputs are the fixed sample in epochs.

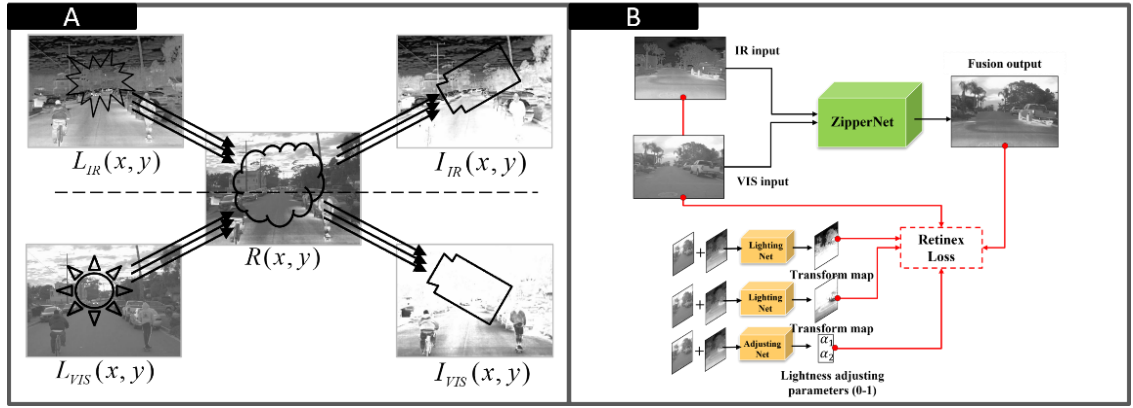

Fig. 1. (A) Retinex theory; (B) DRF framework.

*2.2 Architectures*

In the DRF framework, the fused image is generated by ZipperNet, a dual-path feature switching and skipping network. LightingNet, the single-path feature-skipping networks, generates the transform maps of IR and VIS images. AdjustingNet is a single-path feature skipping network similar to LightingNet, except that only the central (one or two) pixels are used as the overall lightness adjusting parameter(s). Essentially, all of these networks are encoder-decoder architectures, which perform significantly well in generative tasks, especially image-to-image translation [18-21].

Fig. 2(A) reveals the ZipperNet architecture. ZipperNet can handle various inputs and fuse their deep features. The inputs are the low-resolution IR and VIS images of the same size. The backbone of the ZipperNet is composed of 10 encoder blocks and 5 decoder blocks, and they are symmetric with the mirror line. Each encoder block is composed of two 1-stride 3×3 convolution layers with reflection padding to extract features, a batch normalization layer to prevent gradient explosions and vanishes, a leaky-ReLU activating layer, a 1-stride 3×3 convolution layer, a 2-stride 3×3 convolution layer for down-sampling, a batch normalization layer, and a leaky-ReLU layer, successively. These encoder blocks construct the dual-path encoder. To better fuse the deep features, features in different odd-even layers are added onto the same-depth layer in the other path. Identically, each decoder block is composed of a ×2 bilinear layer for up-sampling, a batch normalization layer, two 1-stride 3×3 convolution layers with reflection padding, a batch normalization layer, a leaky-ReLU layer, two 1-stride 3×3 convolution layers with reflection padding, a batch normalization layer, and a leaky-ReLU layer, successively. Furthermore, to fuse the features between the encoder and decoder parts, the skip connections, which are symmetric with the mirror line, are employed to concatenate the features. At the end of the ZipperNet, a sigmoid activation function is adopted to format the value range of the output. The number of convolution kernels is 128 in each encoder block and 128 in each decoder block.

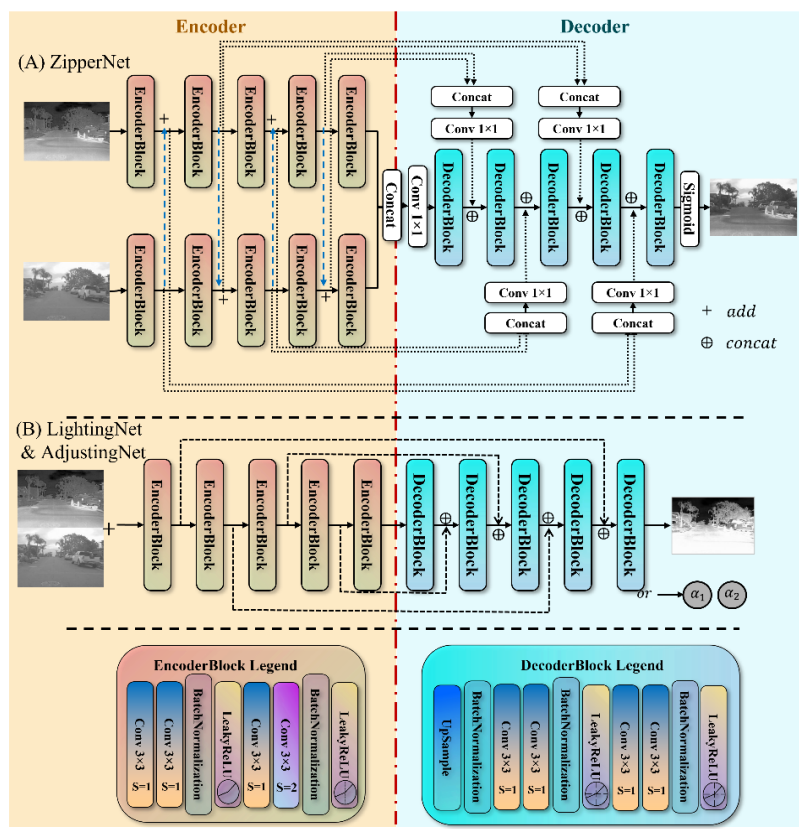

Fig. 2. DRF architectures. (A) ZipperNet; (B) LightingNet and AdjustingNet.

In addition, LightingNet and AdjustingNet have the same architecture as illustrated in Fig. 2(B). LightingNet and AdjustingNet are composed of 5 encoder blocks, 5 decoder blocks, and skipping connections, which are symmetric with the mirror line. LightingNet can generate the transform maps. Different from exploiting the whole output map of LightingNet, AdjustingNet only exploits the central pixels of the output map as the weighted parameters to adjust the lightness of the generated image. LightingNet and AdjustingNet have 8, 16, 32, 64, and 128 convolution kernels in each encoder block and 128, 64, 32, 16, and 8 convolution kernels in each decoder block.

*2.3 Adaptive Fusion Loss Functions*

The loss functions include Retinex loss, joint gradient loss, and lock losses. Among them, the Retinex loss is the core of the DRF. The original design of Retinex loss is demonstrated in Eq. (1), where $H$ and $W$ are the height and width of the image, $R$ is the generated image of ZipperNet, $L^1$ and $L^2$ are the generated images of LightingNet, $I^1$ and $I^2$ are the input IR and VIS images, and $i$ and $j$ are the indexes of pixels. The L1 norm is used here to outperform the L2 norm in pixel-to-pixel tasks [22].

$$\mathcal{L}^*_{\text{Retinex}} = \frac{1}{H \cdot W} \sum_i \sum_j (|R_{i,j} \cdot L^1_{i,j} - I^1_{i,j}| + |R_{i,j} \cdot L^2_{i,j} - I^2_{i,j}|) \tag{1}$$

However, the random initialization generated results of $R$ and $L$ have a high probability of existing zero values, and any zero value in them will cause instability. Meanwhile, untreated maps $R$ and $L$ have large lightness dynamic ranges which can be hardly restricted. To solve these problems, we employed log on Eq. (1) to transfer multiplication into sum as shown in Eq. (2) and reduce the lightness dynamic ranges of $R$ and $L$, respectively.

$$\log I(x, y) = \log R(x, y) + \log L(x, y) \tag{2}$$

Based on Eq. (2), the optimized design of Retinex loss can be obtained as Eq. (3).

$$\mathcal{L}_{\text{Retinex}} = \frac{1}{H \cdot W} \sum_i \sum_j \Big[ \big| \alpha_1 \cdot (\log|R_{i,j}| + c) + \log|L^1_{i,j}| + c) \\ - \log|I^1_{i,j}| + c\big| + \big| \alpha_2 \cdot (\log|R_{i,j}| + c) + \log|L^2_{i,j}| + c) \\ - \log|I^2_{i,j}| + c \big| \Big] \tag{3}$$

Additionally, to avoid zero and negative values in the logarithm, we introduced a small bias $c$ (=$10^{-7}$) and used an absolute value operation before applying the logarithm. Although the optimized Retinex loss can successfully work, the overall fused image is darker than the inputs because of fixed fusion proportion of lightening maps $L$. Additional $α_1$ and $α_2$ ($0<α_1, α_2<1$) as weighted learnable parameters generated by AdjustingNet are introduced to improve the visual lightness perception of fused images.

High-quality IR and VIS fusion requires high-frequency information. Thus, the maximal gradient map between inputs can almost represent the fused gradient map. The joint gradient loss in Eq. (4) is designed to force the network to focus on high-frequency information.

$$\mathcal{L}_{joint}^{grad} = \frac{1}{H \cdot W} \sum_i \sum_j \left| \nabla^2 R_{i,j} - \max(\nabla^2 I_{i,j}^1, \nabla^2 I_{i,j}^2) \right| \quad (4)$$

Furthermore, there is no limit on transform maps $L$ and lightness weighted parameters $\alpha_1$ and $\alpha_2$ during iterations. Here, the $L$ and $\alpha$ lock losses in Eqs. (5) and (6) are designed to limit the value range of $L$, $\alpha_1$, and $\alpha_2$.

$$\mathcal{L}_{lock}^{L} = \frac{1}{H \cdot W} \sum_i \sum_j (\left| L_{i,j}^1 - 1 \right| + \left| L_{i,j}^2 - 1 \right|) \quad (5)$$

$$\mathcal{L}_{lock}^{\alpha} = \left| \alpha_1 - 0.5 \right| + \left| \alpha_2 - 0.5 \right| \quad (6)$$

Based on the Retinex loss, another lightness lock loss in Eq. (7) is designed to keep the lightness of the generated fused image approaching the inputs, where $\bar{R}$ indicates the mean pixel value.

$$\mathcal{L}_{lock}^{mean} = \frac{1}{H \cdot W} \sum_i \sum_j (\bar{R} - \frac{\bar{I}^1 + \bar{I}^2}{2}) \quad (7)$$

Therefore, the total loss is expressed in Eq. (8), and the values of $\lambda_1$, $\lambda_2$, $\lambda_3$, $\lambda_4$, and $\lambda_5$ are 1, 0.2±0.1, 0.25, 0.25, and 1 according to our experience.

$$\begin{aligned}\mathcal{L}_{total} &= \lambda_1 \cdot \mathcal{L}_{Retinex} + \lambda_2 \cdot \mathcal{L}_{joint}^{grad} \\ &+ \lambda_3 \cdot \mathcal{L}_{lock}^{L} + \lambda_4 \cdot \mathcal{L}_{lock}^{\alpha} + \lambda_5 \cdot \mathcal{L}_{lock}^{mean}\end{aligned} \quad (8)$$

In summary, the core idea of DRF is to disentangle the input(s) into components which are the essential parts for a specific physical model via generative networks, and then extract the image low-level statistic prior using network structure. Moreover, to implement physics driven unsupervised learning, the loss functions are designed based on the Retinex theory. Therefore, these loss functions can combine the component priors into a reconstruction closed loop. This process is mapping a specific physical model into a deep learning version via networks by exploiting the self-similarity of the source input(s), low-level statistics prior of source input(s), and handcrafted prior of networks to obtain good performance in IR and VIS fusion.

## 3. Verifications

To verify the performance of the proposed DRF, we qualitatively compared it with the state-of-the-art methods including DDcGAN [15], DenseFuse [23], IFCNN [11], RFN-Nest [24], NestFuse [25], and U2Fusion [26] using the FLIR [27], TNO [28], and VIFB [29] datasets. Different from data-driven methods, our physics driven DRF only uses two fixed IR and VIS inputs in each epoch rather than the training dataset. In addition, most dataset images are in gray both in IR and VIS, and lightness is more important than colors. Therefore, we transferred all inputs into the gray mode in preprocessing. The networks were trained with a learning rate of $10^{-3}$ and iterated for 5,000 epochs. This verification was implemented on the RTX 3060 @6GB GPU and AMD Ryzen 9 @3.3GHz CPU.

Figs. 3(A)-3(C) reveals three representative results using the FLIR [27], TNO [28], and VIFB [29] datasets, respectively. Additionally, Figs. 3(D) and 3(E) exhibit the results in the day and night scenes using the FLIR dataset. In general, an ideal IR and VIS image fusion should be the integration of high-temperature targets in the IR image and a low-temperature background in the VIS image. According to the fusion results in Fig. 3, DRF can preserve abundant VIS texture details as well as high-contrast IR information, thus adaptively balancing the IR and VIS information during image fusion. Moreover, DRF can well provide distinctions between day and night scenes. Therefore, our proposed physics driven DRF has the best performance in IR and VIS image fusion.

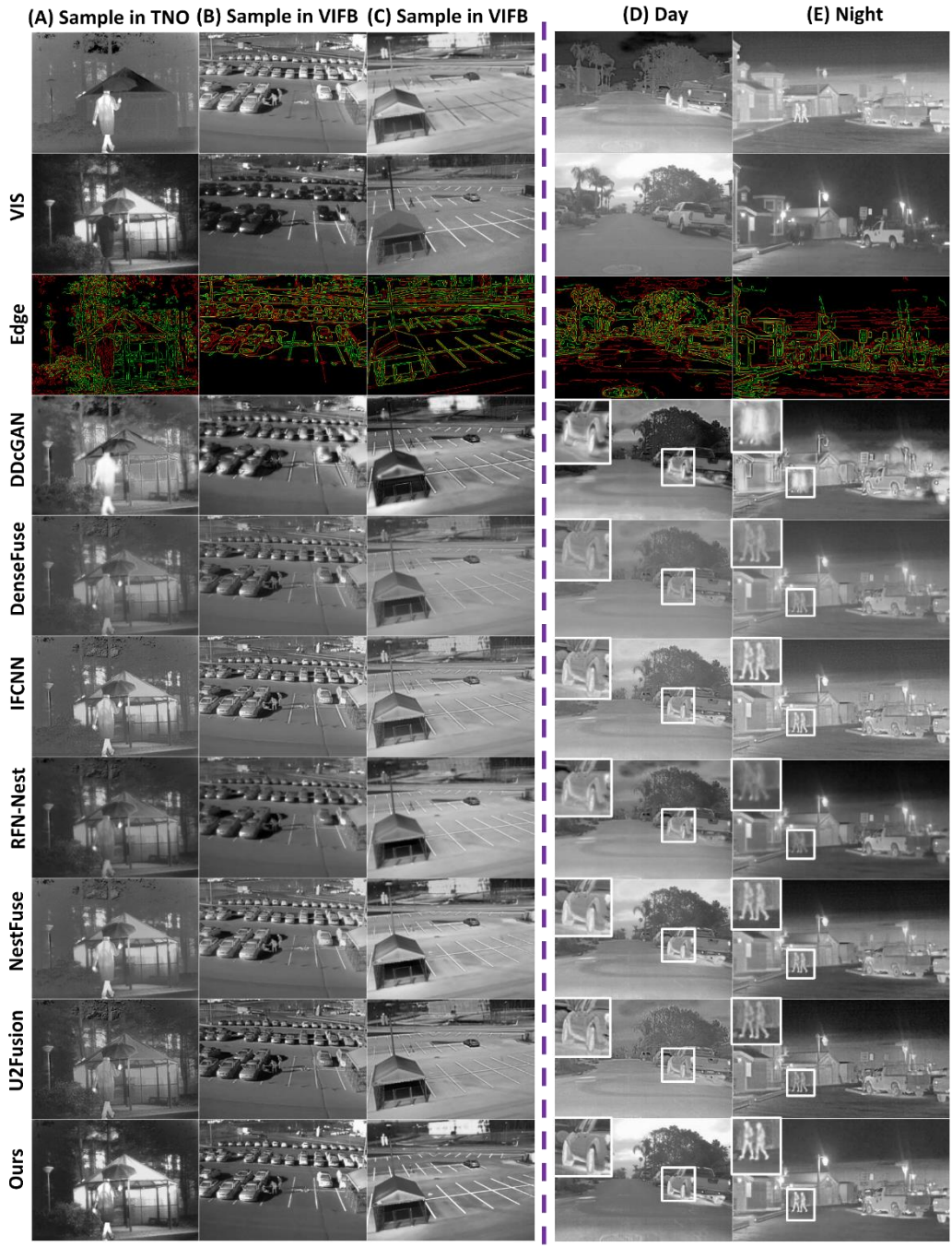

Fig. 3. Comparisons of our approach against multiple state-of-the-art fusion methods (DDcGAN, DenseFuse, IFCNN, RFN-Nest, NestFuse, and U2Fusion) using the FLIR, TNO and VIFB datasets.

**Table 1. Quantitative comparison of our approach against state-of-the-art learning-based methods according to MG, CEN, EI, and SF**[a]

|  | Dataset | DDcGAN | DenseFuse | IFCNN | RFN-Nest | NestFuse | U2Fusion | DRF |
|---|---|---|---|---|---|---|---|---|
| Training Data | ---- | √ | √ | √ | √ | √ | √ | × |
| MG ↑ | FLIR | 2.8224 | 2.0512 | 3.6517 | 1.9033 | 3.0943 | 3.7888 | **4.0045** |
| | TNO | 3.5773 | 1.8624 | 3.4286 | 1.6621 | 2.6127 | 3.2500 | **4.1126** |
| | VIFB | 3.6503 | 2.0003 | 3.6443 | 2.0109 | 2.6237 | 3.2992 | **4.1345** |
| CEN ↓ | FLIR | 1.0674 | 1.0572 | 1.1840 | 1.3543 | 1.1178 | 1.1254 | **0.7810** |
| | TNO | 1.3941 | 1.3038 | 1.5360 | 1.5506 | 1.5001 | 1.4970 | **1.0108** |
| | VIFB | 6.9814 | 4.6489 | 5.2389 | 6.5581 | 7.2613 | 6.6250 | **3.6985** |
| EI ↑ | FLIR | 29.6982 | 21.1764 | 37.6496 | 20.3695 | 32.1763 | 38.4379 | **38.9259** |
| | TNO | 35.0284 | 18.2814 | 33.8043 | 17.5379 | 26.0180 | 33.1625 | **35.4144** |
| | VIFB | 37.5498 | 20.5047 | 37.2774 | 21.5663 | 26.9138 | 34.5566 | **39.6117** |
| SF ↑ | FLIR | 6.2729 | 4.3099 | 7.6609 | 3.9975 | 6.9372 | 7.7557 | **9.1759** |
| | TNO | 6.5438 | 3.6571 | 6.6726 | 3.2651 | 5.3974 | 6.1693 | **8.2946** |
| | VIFB | 8.5957 | 5.0724 | 9.2192 | 4.9092 | 7.1645 | 7.9476 | **11.3476** |

[a] Bold marks the best performance.

We further quantitatively compared our designed DRF with DDcGAN, DenseFuse, IFCNN, RFN-Nest, NestFuse, and U2Fusion, and the results are listed in Table 1 and Fig. 4. The evaluation coefficients include mean gradient (MG), cross entropy (CEN), edge intensity (EI), and spatial frequency (SF) [30]. MG quantifies the high-frequency contents in the fused image, CEN estimates the similarity of image information distribution between source images and fused images, EI reflects the edge intensity calculated by the Sobel operator, and SF describes the richness of the texture details. A high-quality fused image should have high MG, EI, and SF but low CEN. According to Table 1, our proposed DRF has the best performance due to the highest statistical MG, EI, and SF and the lowest statistical CEN in all the conditions.

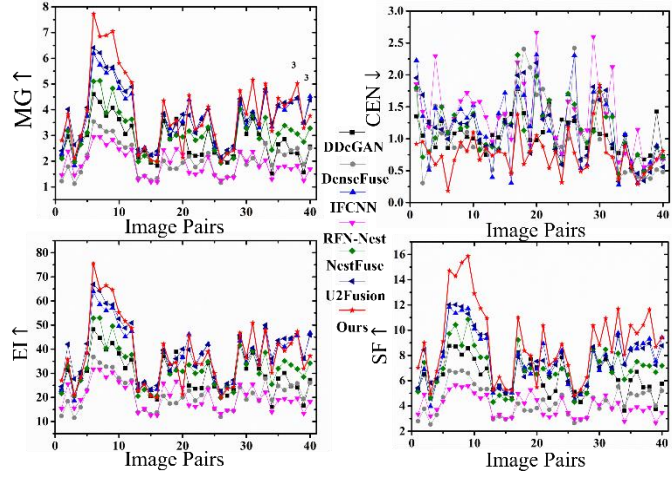

Fig. 4 Quantitative comparison of our approach against state-of-the-art learning-based methods according to MG, CEN, EI, and SF

According to the above qualitative and quantitative comparisons using public datasets, DRF has the best IR and VIS fusion performance. Additionally, DRF can provide distinctions between day and night scenes and preserve abundant VIS texture details and high-contrast IR information. As a result, the proposed DRF outperforms previous works in IR and VIS fusion.

**4. Applications**

Besides the above verifications using the public datasets, we also performed validation*s* of DRF in the practical scenes, which consist of complex indoor instruments with shielding relations and local heat dissipation as shown in Fig. 5. The imaging system consists of a VIS camera (iPhone 13) for VIS image recording as Figs. 5(A1) and 4(A2), and another IR camera (FLIR A655sc) for IR image recording as Figs. 5(B1) and 5(B2). Additionally, Figs. 5(C1) and 5(C2) reveal the IR images in pseudo color. Since these scenes were captured by two different cameras, both the IR and VIS images of the same scene were preprocessed by computational correction and registration before fusion. Figs. 5(D)-5(J) show the IR and VIS fused results corresponding to the DDcGAN, DenseFuse, IFCNN, NestFuse, RFN-Nest, U2Fusion, and our designed DRF, respectively. These results clearly demonstrate that the fusion results obtained by DRF not only well preserve the abundant VIS texture details but also maintain the high-contrast IR information, especially shown in zoomed-in fields of interest. Besides, quantitative comparison simply focusing on the SF also proves that our designed DRF has the best IR and VIS fusion performance.

Significantly, though all the above comparisons were trained with large datasets, our proposed DRF without any external data still had the best fusion performance.

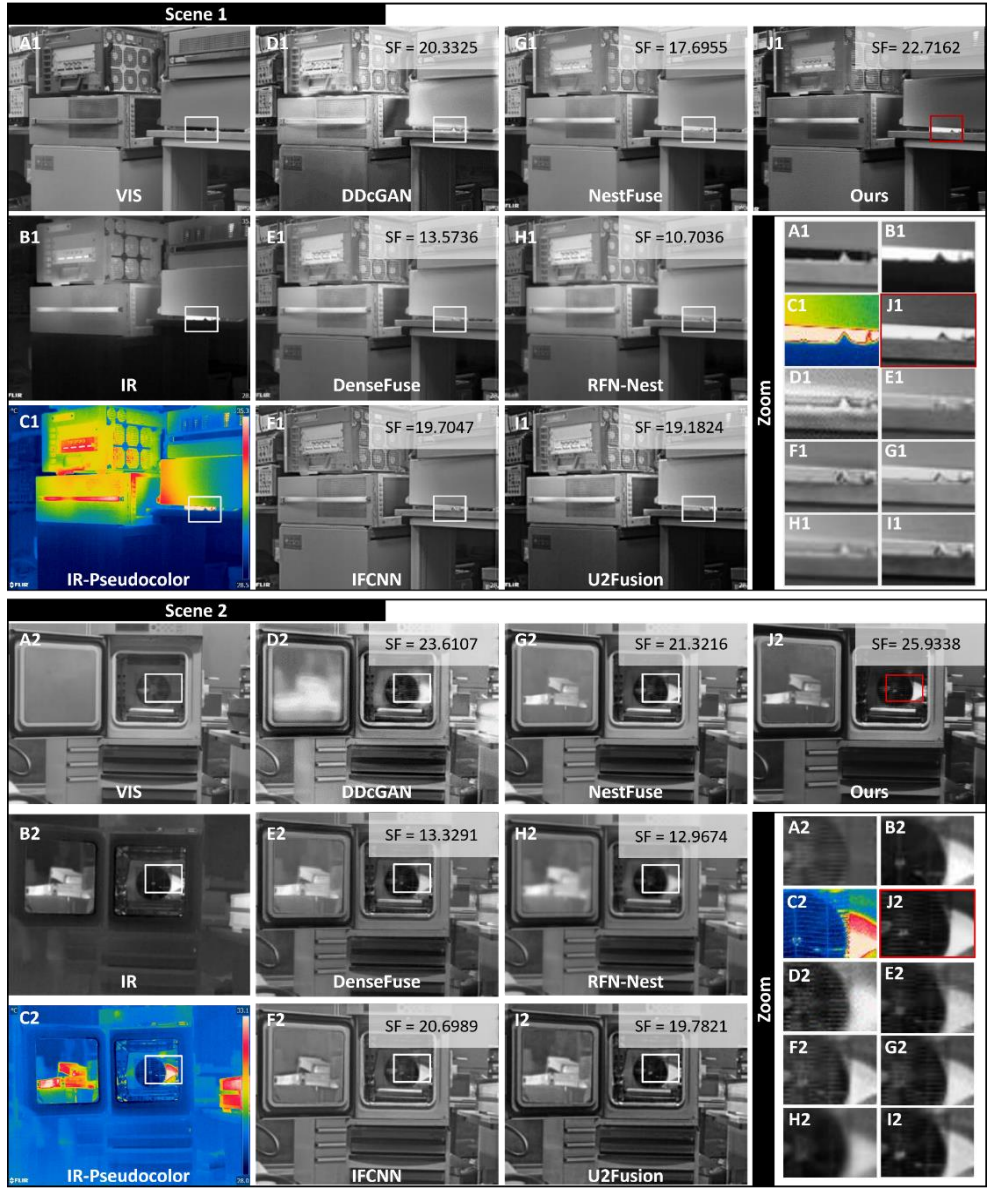

Fig. 5. Comparisons of our approach against the state-of-the-art fusion methods in the practical applications. (A) VIS images; (B) IR images; (C) IR images in pseudo color; (D)-(J) IR and VIS fused results corresponding to the (D) DDcGAN, (E) DenseFuse, (F) IFCNN, (G) NestFuse, (H) RFN-Nest, (I) U2Fusion, and (J) our designed DRF. (1) and (2) represent two different scenes.

## 5. Discussion

Firstly, to verify the effect of our proposed loss functions, we implemented visual ablation experiments in which DRF was trained with different loss functions. Especially in Fig. 6, we discussed the use of each component in the loss function described in Eq. (8). Figs. 6(A) and 6(B) reveal the IR and VIS images in the FLIR dataset. Figs. 6(C)-6(G) list the fusion results corresponding to the loss function without the learnable adjusting parameters $α_1$ or $α_2$, loss function without the lock losses described in Eqs. (5), (6), and (7), loss function without the joint gradient loss described in Eq. (4), loss function in the multiplication form, and our designed loss function, respectively. According to these comparisons, the learnable adjusting parameters $α_1$ and $α_2$ can effectively equalize the histogram, the joint gradient loss can retain more high-frequency details, the lock losses can constrain the learnable adjusting parameters, reflection maps, and lightness level, and the loss functions in the logarithm form can adaptively balance the lightness. Additionally, as shown in Fig. 6(H), with the limitation of lock losses in Eqs. (6) and (7), the learnable adjusting parameters $α_1$ and $α_2$ can adaptively converge to 0-1 according to the lightness level of source inputs. All these comparisons demonstrate our designed logarithm-formed loss functions considering learnable adjusting parameters, joint gradient loss, and lock losses performed well in this DRF task.

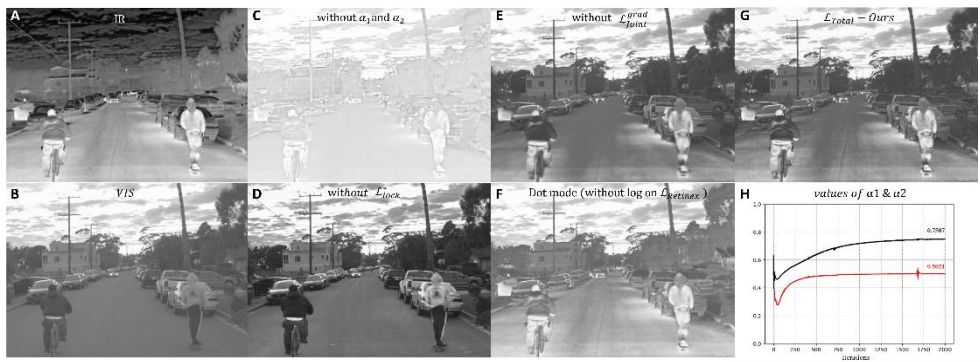

Fig. 6. Study on the effect of the loss functions in DRF. (A) IR and (C) VIS images in the FLIR dataset; (C)-(G) IR and VIS fused images corresponding to (C) the loss function without the learnable adjusting parameters α1 or α2, (D) loss function without the lock losses, (E) loss function without the joint gradient loss, (F) loss function in the multiplication form, and (G) our designed loss function, respectively; (H) Convergence of learnable adjusting parameters $α_1$ and $α_2$.

Secondly, DRF is resistant to noise. In Fig. 7, Figs. 7(A) and 7(B) reveal the IR and VIS images in the TNO dataset, and Figs. 7(C), 7(D), 7(E), and 7(F) show the IR and VIS fused images using DDcGAN,

NestFuse, U2Fusion, and DRF, respectively. We used the method [31] blindly estimate the noise level of these images listed in Fig. 7(G). The result obtained by DRF has the lowest noise level, proving DRF has good noise resistance capability.

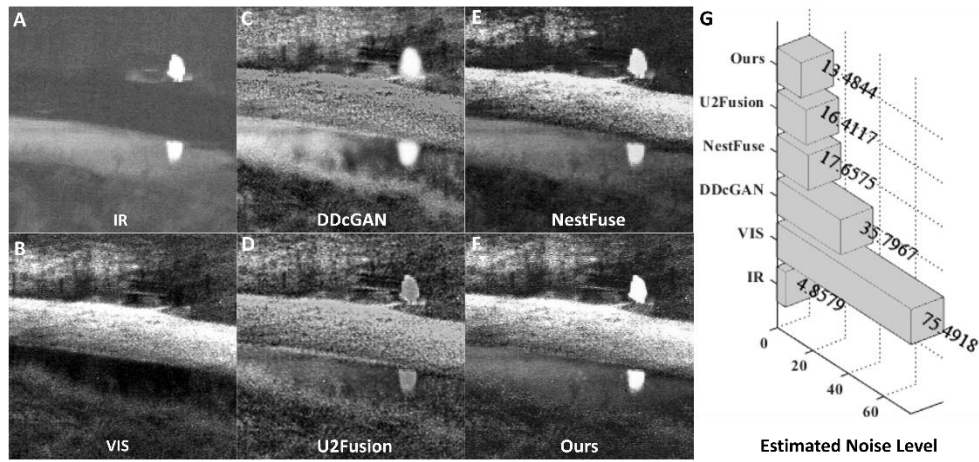

Fig. 7. Study on the noise resistance of DRF. (A) IR and (B) VIS images in the TNO dataset; (C)-(F) IR and VIS fused images using (C) DDcGAN, (D) NestFuse, (E) U2Fusion, and (F) DRF, respectively; (G) Estimated noise level of these images in (A)-(E).

Thirdly, we studied the adaptive capability of DRF by adjusting the levels of IR images. As shown in Fig. 8(A), the levels value of the original IR image is 1.00, and it varies from 0.01 to 2.00 via manual adjusting. While the VIS image levels remain constant also shown in Fig. 8(B). Results obtained via DDcGAN, DenseFuse, IFCNN, RFN-Nest, NestFuse, and U2Fusion by fusing VIS and IR images at different levels as (1) 2.00, (2) 1.00, (3) 0.50, (4) 0.30, (5) 0.01 are listed in Figs. 8(C)-8(I). As the levels of IR image decrease, the IR images gradually darken, and the fusion results obtained by DenseFuse, IFCNN, RFN-Nest, NestFuse, and U2Fusion significantly become dark in Figs. 8(D)-8(H). While as the levels of IR image increase, the image contrast of the IR images gradually reduces, and the fusion results obtained by DenseFuse, IFCNN, RFN-Nest, NestFuse, and U2Fusion also demonstrate lower contrast in Figs. 8(D)-8(H). Though DDcGAN performs with satisfactory adaptive capability in Fig. 8(C), its fusion quality is still poor. Our proposed DRF always provides high-quality fused images with consistent image intensity and contrast even in different level conditions in Fig. 8(I). Therefore, our proposed DRF can adaptively fuse IR and VIS images, thus remarkably improving the fusion robustness.

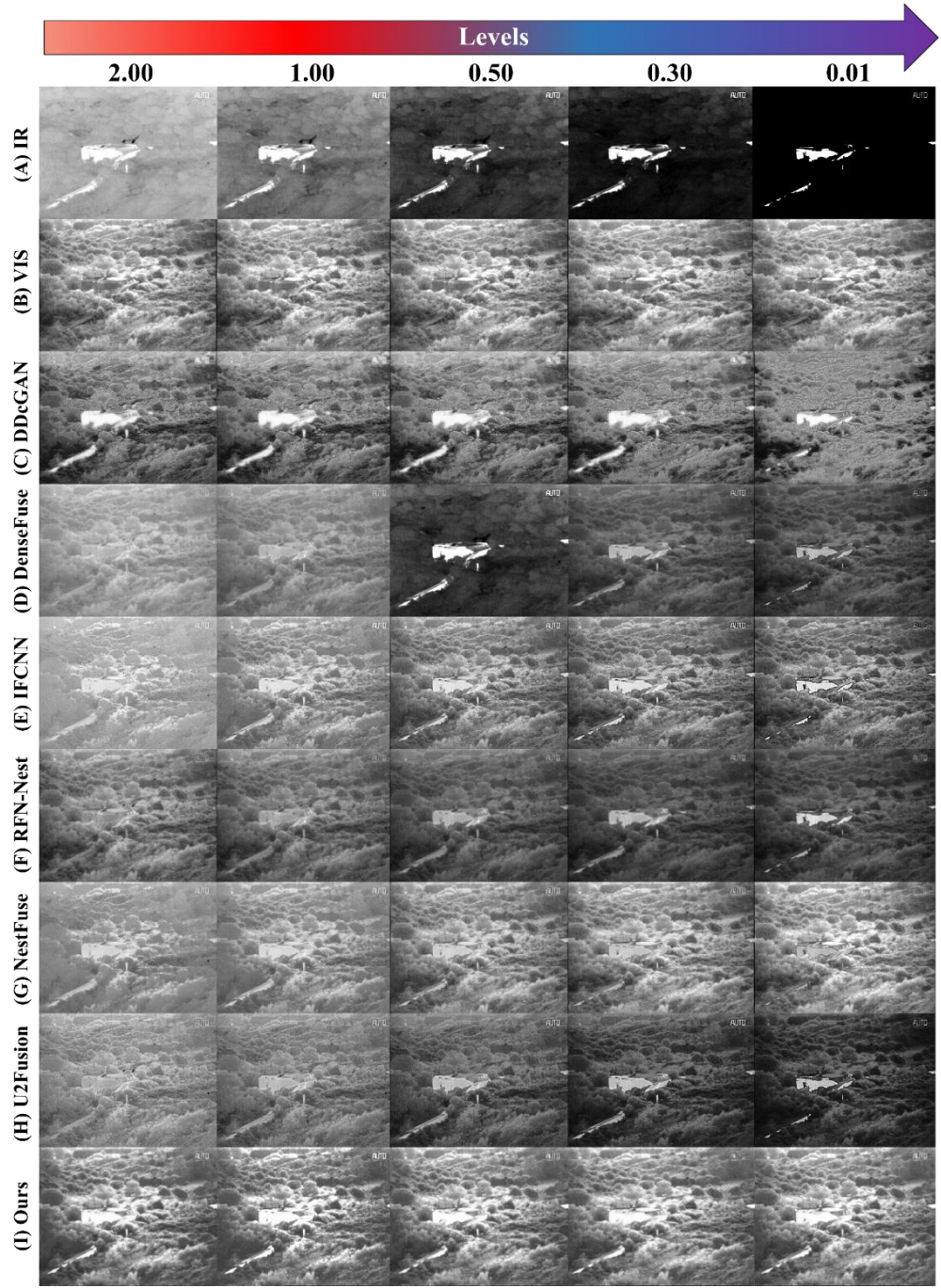

Fig. 8. Study on the adaptive capability of DRF. Comparisons of our approach against multiple state-of-the-art fusion methods (DDcGAN, DenseFuse, IFCNN, RFN-Nest, NestFuse, and U2Fusion) with different IR image levels ranging from 0.01 to 2.

## 6. Conclusion

In this study, we propose physics driven DRF as a dataset-free self-supervised disentangled learning method for adaptive IR and VIS image fusion. In this proposed DRF, generative networks ZipperNet,

LightingNet and AdjustingNet, and Retinex theory-based adaptive fusion loss functions are designed for high-performance image fusion using a unified, dataset-free, and self-supervised model. Compared to many state-of-the-art methods using public datasets and in practical applications, DRF can provide distinctions between day and night scenes and preserve abundant VIS texture details and high-contrast IR information. In addition, DRF can adaptively balance IR and VIS information and has good noise immunity. Therefore, DRF can be a promising tool in multimodal imaging and computational photography.

**Acknowledgement**

This work is supported by National Natural Science Foundation of China (62105196, 61705092), Natural Science Foundation of Jiangsu Province of China (BK20170194), and Shanghai Sailing Program (17YF1407000).

**Author contributions**

**Yuanjie Gu:** Conceptualization, Methodology, Software, Validation, Formal analysis, Investigation, Data Curation, Visualization, Writing – Original Draft, Writing - Review & Editing. **Zhibo Xiao:** Software, Formal analysis, Investigation, Writing - Review & Editing. **Yinghan Guan:** Software, Formal analysis, Investigation, Writing - Review & Editing. **Haoran Dai:** Software, Formal analysis, Investigation, Writing - Review & Editing. **Cheng Liu:** Methodology, Supervision, Writing - Review & Editing. **Liang Xue:** Methodology, Writing - Original Draft, Writing - Review & Editing, Supervision, Project administration, Funding acquisition. **Shouyu Wang:** Conceptualization, Methodology, Writing - Original Draft, Writing - Review & Editing, Supervision, Project administration, Funding acquisition.

**Declaration of interests**

The authors declare that they have no known competing financial interests or personal relationships that could have appeared to influence the work reported in this paper.